\documentclass[letterpaper]{article} 
\usepackage{aaai24}  
\pdfoutput=1
\usepackage{times}  
\usepackage{helvet}  
\usepackage{courier}  
\usepackage[hyphens]{url}  
\usepackage{graphicx} 
\usepackage{natbib}  
\usepackage{caption} 
\usepackage{algorithm}
\usepackage{algorithmic}
\usepackage{subfig}
\usepackage{newfloat}
\usepackage{listings}
\usepackage{multirow}
\usepackage{amsthm}
\usepackage{amsmath}
\usepackage{amssymb}
\usepackage{enumitem}
\usepackage{tabularx}
\newcommand{\myparagraph}[1]{\vspace{0.05\baselineskip}\noindent{\textbf{#1.}}~}
\DeclareCaptionStyle{ruled}{labelfont=normalfont,labelsep=colon,strut=off} 
\floatstyle{ruled}
\newfloat{listing}{tb}{lst}{}
\floatname{listing}{Listing}
\title{Learning to Prompt Knowledge Transfer for Open-World Continual Learning}
\author{
    Yujie Li\textsuperscript{\rm1 \rm3}, 
    Xin Yang\textsuperscript{\rm1 \rm2}\thanks{Corresponding author},
    Hao Wang\textsuperscript{\rm4},
    Xiangkun Wang\textsuperscript{\rm1 \rm2},
    Tianrui Li\textsuperscript{\rm5}
}
\affiliations{
    \textsuperscript{\rm 1}Complex Laboratory of New Finance and Economics, Southwestern University of Finance and Economics\\
    \textsuperscript{\rm 2}School of Computing and Artificial Intelligence, Southwestern University of Finance and Economics\\
    \textsuperscript{\rm 3}School of Management Science and Engineering, Southwestern University of Finance and Economics\\
    \textsuperscript{\rm 4}School of Computer Science and Engineering,  Nanyang Technological University\\
    \textsuperscript{\rm 5}School of Computing and Artificial Intelligence, Southwest Jiaotong University\\

    \{yujie\_li, 222081202013\}@smail.swufe.edu.cn,
    yangxin@swufe.edu.cn,
    cshaowang@gmail.com,
    trli@swjtu.edu.cn
}

\usepackage{bibentry}

\begin{document}

\maketitle

\begin{abstract}
This paper studies the problem of continual learning in an open-world scenario, referred to as Open-world Continual Learning (OwCL).
OwCL is increasingly rising while it is highly challenging in two-fold: i) learning a sequence of tasks without forgetting knowns in the past, and ii) identifying unknowns (novel objects/classes) in the future.
Existing OwCL methods suffer from the adaptability of task-aware boundaries between knowns and unknowns, and do not consider the mechanism of knowledge transfer.
In this work, we propose Pro-KT, a novel prompt-enhanced knowledge transfer model for OwCL. 
Pro-KT includes two key components: (1) a prompt bank to encode and transfer both task-generic and task-specific knowledge, and (2) a task-aware open-set boundary to identify unknowns in the new tasks.
Experimental results using two real-world datasets demonstrate that the proposed Pro-KT outperforms the state-of-the-art counterparts in both the detection of unknowns and the classification of knowns markedly.~\footnote{Code released at \url{https://github.com/YujieLi42/Pro-KT}.}
\end{abstract}

\section{Introduction}

\begin{figure}[ht]
\centering
\includegraphics[width=\columnwidth]{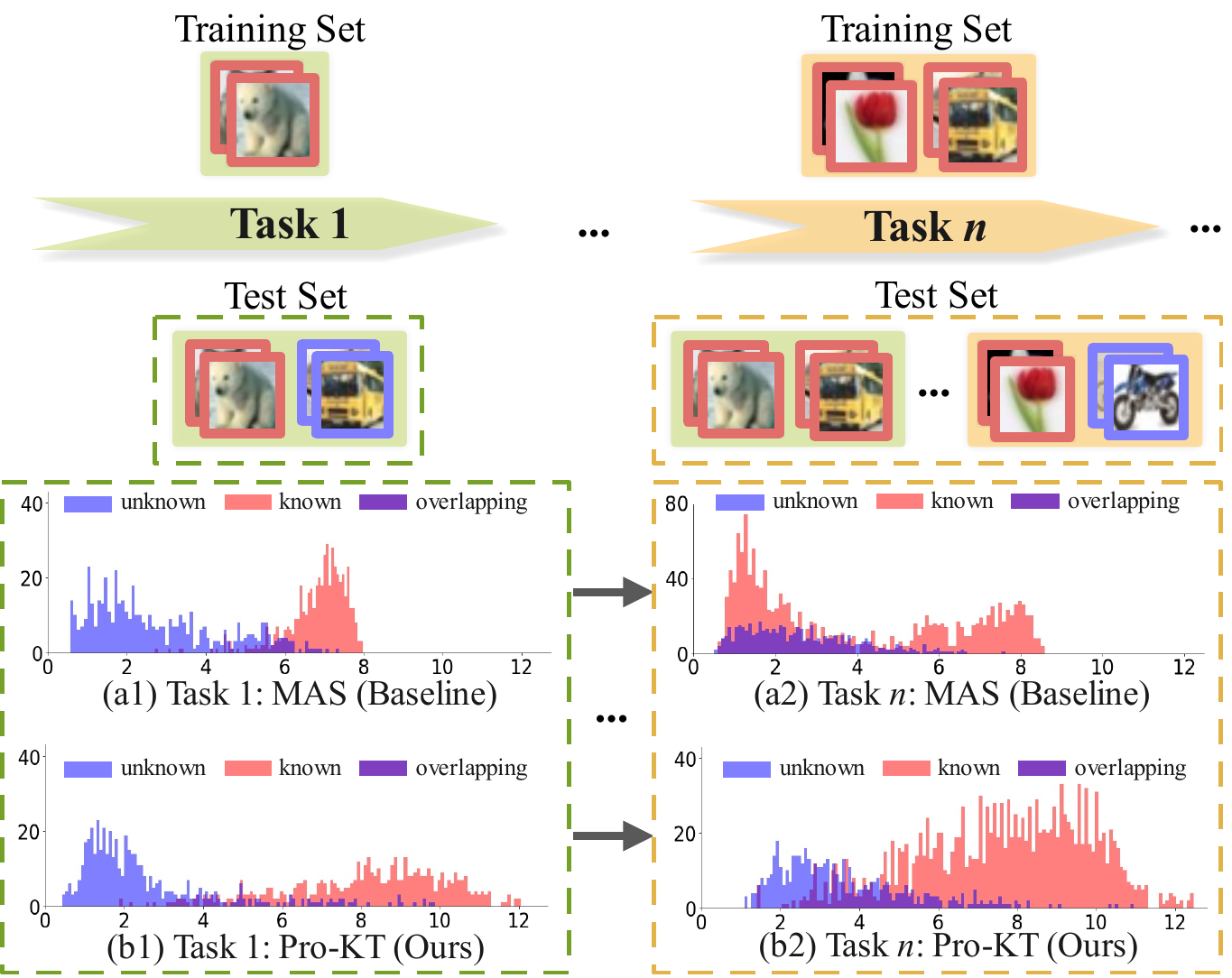}
\caption{An example of our motivation. \textbf{\textit{Top}}: A setting of OwCL that learns a sequence of tasks, where the training data of Task $n$ have new classes versus Task $1$ and the test data of each task may have unknowns versus its training data. \textbf{\textit{Bottom}}: Evaluation results of MAS (a typical CL baseline) and Pro-KT (ours) on Task $1$ and Task $n$, respectively. The evaluation is to evaluate all test data after learning each task. \textit{Note}, the $x$-axis shows the unscaled class-maximum logits of samples, and the $y$-axis shows the number of samples. The area of overlap denotes the performance of partitioning knowns and unknowns (\textit{less is better}).}
\label{fig:intro}
\end{figure}

Continual learning (CL, a.k.a lifelong learning) is a new paradigm of machine learning that typically learns a large number of tasks sequentially without forgetting knowledge gained in the past and uses the knowledge to help new task learning \cite{wang2019forward,ke2020continual}. This paper concerns the problem of CL for a sequence of classification tasks, where each task involves a set of classes to classify. To date, most existing methods, even the most recent counterparts such as \cite{guo2023dealing,smith2023coda}, assume that the sets of training classes and test classes are the same in the past or future tasks, referred to as \textit{closed-world assumption} \cite{fei2016breaking}. However, the closed-world assumption is often invalid in practice as the real world is an open environment where it frequently happens that some tasks contain new classes that are unseen in past tasks. 
For example, the previous task had only learned cats and dogs, but now a new task is to classify a set of birds.
In this work, we study CL with \textit{open-world assumption}, where test data include unknowns or novel classes.

Continual learning in an open world or simply \textit{Open-world Continual Learning} (OwCL) is appealing yet challenging in recent years. Fig.~\ref{fig:intro} illustrates an example of our motivation. Specifically, the scenario refers to an OwCL setting, assuming the training data of Task~$n$ have new classes (i.e., \textit{rose} and \textit{bus}) and both the test data of Task~$1$ and Task~$n$ have novel objects (e.g., \textit{bus} in Task~$1$ and \textit{motorbike} in Task~$n$) that are not seen in the training data. As shown in Fig.~\ref{fig:intro} (a1) and (b1), both MAS \cite{aljundi2018memory} (a typical CL baseline) and our method can distinguish the unknowns from knowns in Task~$1$ (\textit{the first task}). However, as shown in Fig.~\ref{fig:intro}(a2) and (b2), MAS fails to exploit previous unknowns (e.g., $bus$ in Task~$1$) to help Task~$n$, leading to the issue of \emph{performance degradation} on Task~$n$. Our method can separate the knowns from unknowns for Task~$n$. Our main motivation here is to learn knowledge for both knowns and unknowns from previous tasks and later use the knowledge to formulate an open-set boundary between knowns and unknowns for new tasks. 

In this work, we aim at designing an OwCL model that accumulates the knowledge gained in the past and uses the knowledge to help divide the test data into knowns and unknowns/opens, especially for enlarging the open-set boundary between knowns and unknowns. After learning each new task, the newly seen classes are subsequently treated as knowns for new tasks. Our main ideas are \textit{knowledge transfer} and \textit{prompt learning}. However, there are two major challenges. (1) \textit{Knowledge Stability}: When a new task is largely dissimilar from the previous tasks, the newly acquired knowledge might contradict the existing knowledge. (2) \emph{Knowledge Plasticity}: Once the ground-truth data of unknowns appear in the subsequent tasks, the model needs to update the corresponding unknown samples as knowns.

To address the aforementioned challenges, we propose a novel \textbf{Pro}mpt-enhanced \textbf{K}nowledge \textbf{T}ransfer (Pro-KT) method. Pro-KT delineates an innovation of prompt learning for OwCL, a novel plug-and-play prompt bank for knowledge transfer, and two adaptive threshold selection strategies for determining the open-set boundary. Specifically, to address challenge (1), we create a prompt bank designed to encode knowledge through the use of prompts. These prompts serve as instructions for directing the model in task execution. 
By flexibly selecting prompts from the proposed prompt bank, Pro-KT can facilitate effective knowledge transfer with both task-generic and task-specific knowledge across diverse tasks. 
To address challenge (2), we design two adaptive threshold-selection strategies for determining the open-set detection boundary. Through these strategies, the open-set decision boundary will be updated according to the newly learned knowledge continually so as to handle knowledge plasticity.

Our main contributions are summarized as follows:
\begin{itemize}[leftmargin=*] 
\item This paper delves into knowledge transfer via prompt learning for OwCL. We propose a novel prompt bank for rehearsal-free knowledge transfer that delineates a new paradigm for representing, accumulating and transferring knowledge. 
\item We propose a novel task-aware open-set boundary to distinguish unknowns from knowns, based on two adaptive threshold-selection strategies. Thus, the model is able to accumulate and revise the knowledge learned from unknowns, ensuring knowledge plasticity for OwCL.
\item Extensive experiments on two sets of real-world datasets demonstrate that the proposed Pro-KT outperforms baseline methods markedly on a large range of evaluation tasks. Case study and visualization results further show Pro-KT's effectiveness in tackling OwCL.
\end{itemize}

\section{Related Work}
There are two closely related research topics: \textit{Open-world Continual Learning} and \textit{Prompt Learning}.

\myparagraph{Open-world Continual Learning}
OwCL is a synergistic integration of continual learning \cite{pmlr-v70-zenke17a} and open-world learning \cite{bendale2015towards}. In a nutshell, OwCL aims at learning on the job in the open world with the goal of recognizing unknowns and incrementally learning them without catastrophic forgetting so that a model will become more and more knowledgeable for future learning. There are mainly two OwCL settings: 1) \textit{Task-incremental} OwCL, where the classes (if any) in the tasks may or may not be disjoint, and 2) \textit{Class-incremental} OwCL, where each task has a set of non-overlapping classes. In this work, we concern about the setting of class-incremental OwCL.

\citet{8631004-2019} designed an open-set classifier for identifying new classes based on the extreme value theory and further proposed an ensemble classifier for open-set incremental learning.
\citet{Joseph_2021_CVPR} presented a solution for identifying unknown categories without forgetting learned classes based on contrastive clustering and energy-based unknown identification.
\citet{kim2023open} theoretically demonstrated the necessity of novelty detection for class-incremental continual learning. 
\citet{liu2023ai} introduced a prospective framework, SOLA, to facilitate autonomous continual learning through steps including novelty detection, adapting new tasks on the fly, and incremental learning.

However, the above-mentioned methods do not consider knowledge representation and integration schemes for OwCL.
In addition, there is a lack of effective methods concerning the representation, accumulation, and update of knowledge for unknowns.

\myparagraph{Prompt Learning}
With the development of large-scale pre-trained models, prompt learning \cite{zhou2022learning,khattak2023maple,gu2022ppt,wang2022learning} is an emerging technique that deploys pre-trained models in a parameter-efficient and data-efficient way \cite{liu2023pre}. 
It applies a fixed function to guide a pre-trained model so that the pre-trained model achieves additional instructions to perform downstream tasks \cite{lester2021power,li2021prefix}. 
Owing to the advantages of prompts, prompt-based continual learning methods show strong protection against forgetting by learning insertable prompts rather than modifying or constraining encoder parameters directly \cite{smith2023coda}.

Prompt learning has been investigated recently for continual learning, such as L2P \cite{wang2022learning}, DualPrompt \cite{wang2022dualprompt} and CODA \cite{smith2023coda}. 
However, the existing prompt learning-based methods have certain limitations. For example,
L2P requires a single prompt pool to be updated after every new task, resulting in performance degradation for a long sequence of tasks. 
DualPrompt learns a set of task-specific prompts and a unique global prompt without considering knowledge transfer across tasks. 
CODA includes an end-to-end attention mechanism to decompose prompts, while learning a large number of prompts and increasing the prompt length to achieve effective results.
What's more, existing prompt-based methods follow the closed-world assumption, thereby lacking the capability to identify unknowns within an open-world context.

\begin{figure*}[ht]
\centering
\includegraphics[width=\textwidth]{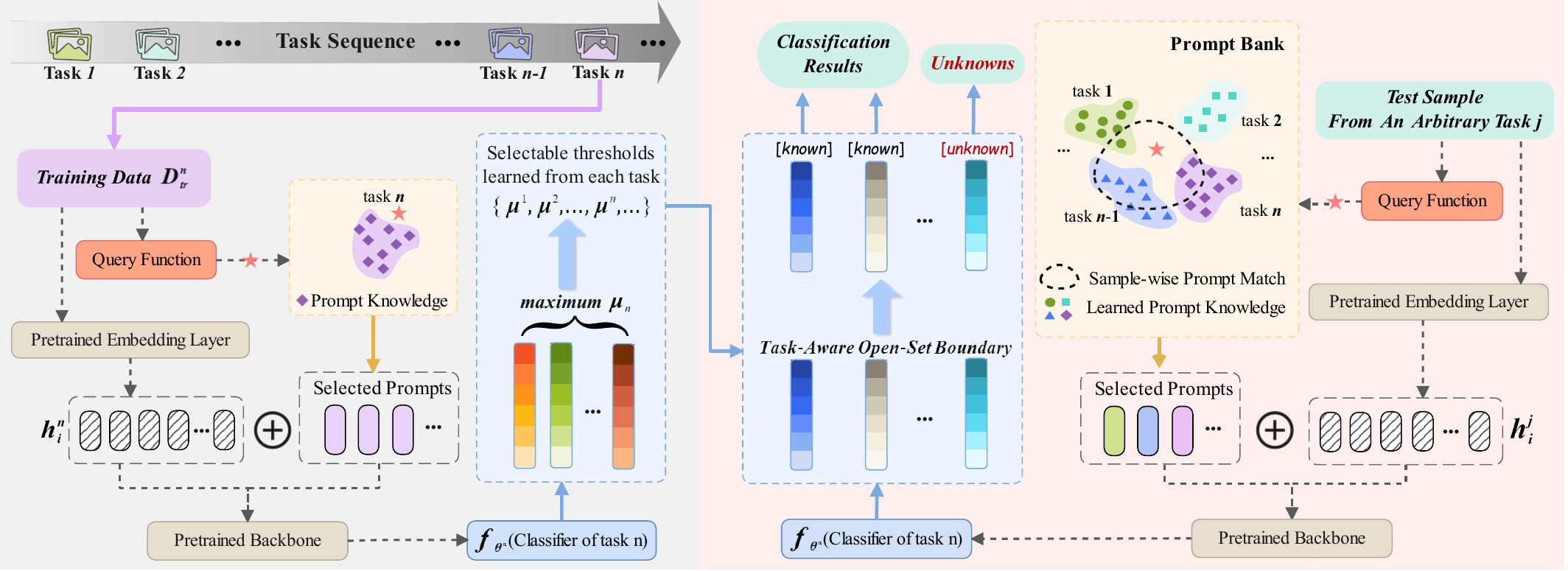}
\caption{Illustration of the overall framework (\textbf{\textit{Left}}: Training phase, and \textbf{\textit{Right}}: Test phase). [Best viewed in color]}
\label{fig:framework}
\end{figure*}

\myparagraph{\textit{Remarks}} As aforementioned, typical continual learning makes the closed-world assumption, which is often invalid in practice as the real world is an open environment that is full of unknowns or novel objects. Although there are several attempts with the open-world assumption, OwCL is still a highly challenging research problem. 
Most existing OwCL methods suffer from misinterpreting, assembling, or revising incorrect knowledge, without focusing on knowledge transfer. 
Our method, Pro-KT, addresses these limitations with prompts to encode knowledge and learns to prompt knowledge transfer for OwCL. Pro-KT buffers knowledge in a more intelligent and succinct rehearsal-free mechanism. 
The details of the proposed method will be clear shortly. 

\section{Methodology}

In this section, we first clarify the research problem and the overall framework. Then we elaborate on two key components of the proposed Pro-KT: (1) \textit{prompt bank}, and (2) \textit{task-aware open-set boundary}. Finally, we conclude the overall objective function. Before going further, let us make a notational convention used throughout the paper. We will use superscripts to denote task identifiers (task IDs) for a sequence of tasks (e.g., $T^1$, ..., $T^n$).

\myparagraph{Problem Statement}
Given a sequence (possibly never-ending) of tasks, our OwCL problem is that during continual learning if a learner encounters any novel objects in the test, the learner should detect them as `unknowns'. Then, suppose the learner approaches the ground truths for these `unknowns' in a new task, the `unknowns' will be evolved as knowns. What's more, the learner should incrementally learn on the job -- namely, \textit{recognizing unknowns and incrementally learning them without catastrophic forgetting for new task learning}. Note that the training data from the previous tasks may be not accessible again for training new tasks. Briefly, we define our problem as follows:

\noindent \textbf{\textit{Problem Definition}}. At any point in time, the learner has learned $n$ tasks. Each task $T^n$ has its training data $D_{tr}^n$. When faced with a new task $T^{n+1}$ with its training data $D_{tr}^{n+1}$, the purpose of our OwCL problem is that the learner 1) detects unknows from new tasks, 2) annotates unknowns as knowns along with time, and 3) classifies previous classes without forgetting.

\myparagraph{Architecture} Fig.~\ref{fig:framework} illustrates the overall framework of the proposed Pro-KT, comprising the \textit{training} and \textit{test} phases.

In the \textit{training,} samples are first projected with a query function. Pro-KT then learns a new set of prompts and accumulates them into the prompt bank. Next, each training sample is enhanced by chosen prompts, and all these prompt-enhanced embeddings are fed into the pre-trained backbone. The prompt-enhanced final representations are then fed into a trainable classifier. After that, we learn an adaptive threshold for the task-aware open-set boundary, based on the softmax entropy of the prompt-enhanced final representation.
During the \textit{testing} phase, after the projection of test samples, prompts are drawn from the prompt bank employing a sample-wise matching mechanism. This mechanism facilitates the transfer of both task-specific and task-generic knowledge. The enriched embeddings are then forwarded to the unified classifier, yielding unscaled logits scores. Pro-KT then detects samples as either $[known]$ or $[unknown]$ based on the task-aware open-set boundary, guided by adaptive thresholds.

\subsection{Prompt Bank}
We propose to use a prompt bank for knowledge transfer. Specifically speaking, we use prompts to encode both task-generic and task-specific knowledge and retain the prompts in a prompt bank. The prompts stored in the prompt bank are later transferred to new tasks for problem-solving, which further assures knowledge stability in OwCL. Given the current task $T^n$ with its training data $D_{tr}^n=\{(\mathbf{X}^n,\mathbf{Y}^n)\}$, the learner achieves a prompt set $(p_1^n,...,p_i^n,...)$ for this task. Then, the learner stores this portion of knowledge together with the knowledge learned from previous tasks into the prompt bank:
\begin{equation}
\begin{split}
    \mathbf{P} & = \bigcap_{n=1}^N (\mathbf{p}^1,...,\mathbf{p}^n,...,\mathbf{p}^N) \\
    & =\{ (p_1^1,p_2^1,...,p_M^1),..., \\
    & \ (p_1^n,p_2^n,...,p_M^n),..., (p_1^N,p_2^N,...,p_M^N) \},
\end{split}
\end{equation}
where $N$ is the total number of tasks, and $M$ is the number of prompts learned from each task. $p_i^j \in \mathbb{R}^{L_p \times D_e \times NM}$ denotes the $i$-th prompt in task $j$, where $L_P$ is the token length, and $D_e$ is the embedding dimension. 
The prompt bank encourages comprehensive knowledge transfer with task-specific knowledge and task-generic knowledge.

However, a critical problem is how to encode and select the most useful prompting knowledge. Inspired by the key-value pair-based query strategy in L2P \cite{wang2022learning}, we propose to associate each prompt as a value with a learnable key. Specifically, given the task $T^n$, we perform a key-query pair for the sample-wise prompt matching:
\begin{equation}
\begin{split}
    (\mathbf{k}^{n},\mathbf{p}^{n}) = (k_1^n,p_1^n),...,(k_M^n,p_M^n),
\end{split}
\end{equation}
where $\mathbf{p}^{n} = \{ p_m^n \}_{M}$ and $\mathbf{k}^{i} = \{ k_m^n \}_{M}$. 
Then, after learning task $T^N$, the key-query pairs stored in the prompt bank can be formulated as:
\begin{equation}
\begin{split}
(\mathbf{K},\mathbf{P})=\{ (\mathbf{k}^{1},\mathbf{p}^{1}),(\mathbf{k}^{2},\mathbf{p}^{2}),...,(\mathbf{k}^{N},\mathbf{p}^{N}) \}.
\end{split}
\end{equation}

Next, we devise a more automatic knowledge transfer protocol via a sample-wise prompt matching mechanism, where the sample itself elects the most valuable/appropriate prompts as auxiliary knowledge from the prompt bank. Given an arbitrary input ${x_i}^n$ from task $T^n$, we first apply the pre-trained projection layer $\mathbf{Q}_{x}$ to map the initial input to a new feature space $\mathbb{R}^{D_e}$:
\begin{equation}\label{equa:query}
 \mathbf{h}_i^n = \mathbf{Q}_{x} \cdot {x_i}^n.
\end{equation}
Then, we apply the pre-trained projection layer as the query function $\mathbf{Q}_{x}$ to lookup the top-K keys by solving the following optimization problem:
\begin{equation}\label{equa:prompt}
    \underset{[1, K]}{\operatorname{argmin}} \sum_{j=1}^{N} dis(\mathbf{h}_i^n, \mathbf{k}^{j}), \text{with }\mathbf{k}^{j} = \{ k_m^j \}_{M}.
\end{equation}
In such a method, the encoded input sample $\mathbf{h}_i^n$ can find top-K keys by searching the top-K minimum distance (we use cosine similarity in this work) between prompt keys without any task identifier. After this, the input embedding $\mathbf{h}_i^n$ is thus extended by the subset $\mathbf{P}_{S_K} \subset \mathbf{P}$ of prompt knowledge from the prompt bank:
\begin{equation}\label{equa:extend}
 \mathbf{h'}_i^n = \mathbf{h}_i^n \oplus  \mathbf{P}_{S_K},
\end{equation}
where $\oplus$ denotes the dimension-wise concatenation. The size of the subset ${P}_{S_K}$ is $K$, which is smaller than $M$. 

Since unknown samples may appear in the test, these enhanced embeddings will then be exploited to learn an adaptive detection boundary to distinguish knowns and unknowns for each task.

\subsection{Task-Aware Open-Set Boundary}
In this section, we elaborate on how to formulate a boundary between the knowns and unknowns and its adaptability to each task during the OwCL process. To address this issue, we devise two adaptive open-set detection threshold selection schemes for determining such a boundary.

In training, the prompting knowledge stored in the prompt bank is learned from the training data of each task with ground-truth labels.
However, it is essential to use the knowledge gained from the knowns to identify unknown samples. 
We propose to learn a threshold on the softmax entropy of the prompt-enhanced embeddings and then utilize the knowledge acquired from the known entities to determine an open-set boundary between knowns and unknowns.

After training task $T^n$, the prompt-enhanced training sample $\mathbf{h'}_i^n$ is fed into a pre-trained backbone $f_{pr}$. The final representation is fed into a trainable classifier $f_{\theta^n}$: 
\begin{equation}\label{equa:prob}
    \emph{prob}_i^{n} =  f_{\theta^n}(f_{pr}(\mathbf{h'}_i^n)), 
\end{equation}
where the $\emph{prob}_i^{n}$ is the maximum value of unscaled softmax entropy of the training sample $\mathbf{h'}_i^n$. 

Then, we determine the open-set detection threshold $\mu$ by computing the mean of all the training samples $\{\emph{prob}_1^{n}, ..., \emph{prob}_i^{n},...\}$ from task $T^n$:
\begin{equation}\label{equa:threshold}
\begin{split}
    \mathbf{\mu}^n=r \cdot \frac{1}{\left|\mathbf{X}^{n}\right|} \sum^{\left|\mathbf{X}^n\right|}_{i=1} \emph{prob}_i^{n}, 
\end{split}
\end{equation}
where ${\left|\mathbf{X}^n\right|}$ denotes the total number of training samples and $r$ denotes the deviation degree.
Subsequently, the obtained open-set detection threshold of task $T^n$
is stored with the previously learned thresholds. In such a manner, the model has the ability to deliver a suitable threshold from $[\mu^1,...,\mu^{n}]$ for the testing phase.

During the test phase, to determine whether a test sample $x_i^j$ from an arbitrary task $T^j$ is known or not, we design two simple but efficient strategies to choose the most appropriate threshold for the open-set detection boundary.

There are two scenarios. The first scenario is that the task identifiers (task IDs) are available in the test. Therefore, the model directly adapts the corresponding threshold via its task ID $j$:
\begin{equation}\label{equa:open_id}
\begin{split}
        x_i^j &: f_{\theta^n}(\mathbf{h'}_i^j) \leq \mu^{j} \rightarrow [unknown],\\
    & \text{otherwise, } \rightarrow [known].
\end{split}
\end{equation}

The second scenario is that the task IDs are unavailable in the test. During OwCL, the model is incrementally learning new tasks, resulting in an accumulation of knowledge in the prompt bank. 
Hence, in this scenario, our model alternatively selects the latest threshold learned from the current task $T^n$ for the open-set detection boundary:
\begin{equation}\label{equa:open}
\begin{split}
        x_i^j &: f_{\theta^n}(\mathbf{h'}_i^j) \leq \mu^n \rightarrow [unknown],\\
    & \text{otherwise, } \rightarrow [known].
\end{split}
\end{equation}

After this, if a sample is detected as an unknown object, the model then annotates the sample with $[unknown]$ for future tasks.
If a sample is detected as one of the $[known]$ objects, the model then classifies the sample with one of the known classes. 

\subsection{Objective Function}
In the training of task $T^n$, the prompt bank first learns $M$ prompts in total with the corresponding keys. 
The prompt bank $(\mathbf{K},\mathbf{P})$ stores all the prompt knowledge learned from $T^1$ to $T^n$. 
Then, the known samples link to top-K prompt keys by Eq.~\eqref{equa:prompt} and extend with auxiliary knowledge by Eq.~\eqref{equa:extend}. 
The open-set detection boundary is also updated through the proposed threshold-setting strategies.
 
Finally, we formally formulate the overall objective function as follows:
\begin{equation}\label{equa:loss}
    \begin{split}
        \underset{\theta^n,(\mathbf{k}^{n},\mathbf{p}^{n})}{\operatorname{min}} \{ \mathcal{L}( f_{\theta^n}(f_{pr}(\mathbf{h'}^n),\mathbf{y}^n)+\\
        \lambda \sum_{i=0}^{M} dis(\mathbf{h}^n, k^{n}_{i}) \},
    \end{split}
\end{equation}
where the $\mathbf{h}^n$ refers to the set of training samples projected embeddings. The first term is the softmax cross-entropy loss, and the second term is a surrogate loss to pull the selected keys closer to the corresponding query features. $\lambda$ is a trade-off parameter to balance the two terms. 
As previously noted, during the training phase of task $T^n$, we exclusively update the prompt set $p^n$, and during the testing, we select prompts from the entire prompt bank $\mathbf{P}$.

\section{Experiments}\label{sec:experiments}
\subsection{Experimental Setup}
\myparagraph{Implementations}
In our experiments, we remove the labels of the classes in the training set of the next task (i.e., $T^{(n+1)}$) and treat them as unknowns in the test set for the current task (i.e., $T^n$). 
Besides, we followed the general setting in the CL community, i.e., randomly shuffled the task order five times for both Split CIFAR100 (with 10 tasks) and 5-datasets (with 5 tasks).
The results presented throughout this manuscript are the mean results of five random shuffles.

\myparagraph{Datasets} We experiment on two commonly-used and publicly-available datasets, namely \textbf{Split CIFAR100} \cite{krizhevsky2009learning} and \textbf{5-datasets} \cite{ebrahimi2020adversarial}. The Split CIFAR100 dataset is sampled from the CIFAR100 by dividing the original CIFAR100 into 10 tasks, where each task contains 10 disjoint classes. Since the 10 tasks are sampled from the same dataset, they exhibit certain similarities. The 5-datasets comprises five distinct image classification datasets: TinyImagenet, NotMNIST, CUB200, Fashion-MNIST and CIFAR10. Each dataset represents an individual task.

\myparagraph{Baselines}
Since OwCL involves continual learning with novelty detection \cite{kim2023open}, we evaluate two tasks: (1) \textit{Unknown Detection} and (2) \textit{Known Classification}. 
\begin{itemize}[leftmargin=*] 
    \item \textbf{Unknown Detection}: We compare Pro-KT against the following seven baseline methods: \emph{OpenMax} \cite{bendale2015towards}, \emph{MSP} \cite{hendrycks2016baseline}, \emph{MCD} \cite{yu2019unsupervised}, \emph{EnergyBased} \cite{liu2020energy}, \emph{Entropy} \cite{chan2021entropy}, \emph{MaxLogits} \cite{basart2022scaling} and \emph{ODIN} \cite{liang2018enhancing}. 
    \item \textbf{Known Classification}: We compare Pro-KT with the following nine continual learning methods: \emph{EWC} \cite{kirkpatrick2017overcoming}, \emph{MAS} \cite{aljundi2018memory}, \emph{DER++} \cite{buzzega2020dark}, \emph{GEM} \cite{lopez2017gradient}, \emph{LwF} \cite{li2017learning}, \emph{HAT} \cite{serra2018overcoming}, \emph{UCL} \cite{ahn2019uncertainty}, \emph{L2P} \cite{wang2022learning} and \emph{DualPrompt} \cite{wang2022dualprompt}. 
\end{itemize}

\myparagraph{Metrics} 
For unknown detection, we use the average area under the curve ($AUC_N$) and average false positive rate ($FPR_N$) \cite{chan2021entropy} as metrics. $AUC_N$ is the average area under the receiver operating characteristic (ROC) curve across all past tasks, providing a comprehensive measure of open detection performance over $N$ tasks. 
$FPR_N$ reports the average error rate of misclassifying unknown samples into known categories. 

For known classification, we employ average final accuracy ($A_N$) and average forgetting rate ($F_N$) \cite{wang2022learning} as metrics. 
$A_N$ is the average final accuracy concerning all past classes over $N$ tasks.
$F_N$ measures the performance drop across $N$ tasks, offering valuable information about plasticity and stability during OwCL.

\subsection{Main Results}
\begin{table}[ht]\small\centering
\begin{tabular}{c|c|cc}
\hline
 Dataset & Methods & $AUC_N$ & $FPR_N$ \\ \hline
\multicolumn{1}{c|}{\multirow{8}{*}{\begin{tabular}[c]{@{}c@{}}Split CIFAR100\\ (10 tasks)\end{tabular}}} & OpenMax & 49.99 & 50.06 \\
\multicolumn{1}{c|}{} & MSP & 50.29 & 52.18 \\
\multicolumn{1}{c|}{} & MCD & 50.17 & 70.37 \\
\multicolumn{1}{c|}{} & EnergyBased & 49.66 & 71.86 \\
\multicolumn{1}{c|}{} & Entropy & 49.56 & 51.17 \\
\multicolumn{1}{c|}{} & MaxLogits & 52.63 & 74.99 \\
\multicolumn{1}{c|}{} & ODIN & 50.81 & 52.64 \\ \cline{2-4} 
\multicolumn{1}{c|}{} & Pro-KT w/o task IDs & \underline{91.01} & \underline{41.31} \\ 
\multicolumn{1}{c|}{} & Pro-KT & \textbf{92.69} & \textbf{39.71} \\ 
\hline
\multirow{8}{*}{\begin{tabular}[c]{@{}c@{}}5-datasets\\ (5tasks)\end{tabular}} & OpenMax & \multicolumn{1}{c}{60.09} & \multicolumn{1}{c}{84.23} \\
 & MSP & \multicolumn{1}{c}{51.22} & \multicolumn{1}{c}{87.96} \\
 & MCD & \multicolumn{1}{c}{50.60} & \multicolumn{1}{c}{98.62} \\
 & EnergyBased & \multicolumn{1}{c}{49.38} & \multicolumn{1}{c}{87.99} \\
 & Entropy & \multicolumn{1}{c}{39.49} & \multicolumn{1}{c}{96.34} \\
 & MaxLogits & \multicolumn{1}{c}{66.78} & \multicolumn{1}{c}{73.12} \\
 & ODIN & \multicolumn{1}{c}{49.63} & \multicolumn{1}{c}{97.91} \\ \cline{2-4} 
 & Pro-KT w/o task IDs & \underline{82.76} & \underline{50.05} \\ 
 & Pro-KT & \textbf{88.60} & \textbf{45.70} \\ 
 \hline 
\end{tabular}
\caption{Results($\%$) regarding unknown detection. We report the results over 10 tasks for Split CIFAR100 (10 classes per task) and 5 tasks for 5-datasets.}
\label{table_1}
\end{table}

\begin{table*}[ht]
\centering
\begin{tabular}{cp{2cm}<{\centering}cp{2cm}<{\centering}cp{2cm}<{\centering}cp{2cm}<{\centering}cp{2cm}<{\centering}cp{2cm}<{\centering}cp{2cm}<{\centering}cp{2cm}<{\centering}}
\hline
\multirow{2}{*}{Backbone} & \multirow{2}{*}{Method} & \multicolumn{3}{c}{\begin{tabular}[c]{@{}c@{}}Split CIFAR100 (10 tasks)\end{tabular}} & \multicolumn{3}{c}{\begin{tabular}[c]{@{}c@{}}5-datasets (5 tasks)\end{tabular}} \\ \cline{3-8} 
 &  & $A_N$($\uparrow$) & $Diff$($\downarrow$) & $F_N$($\downarrow$) & $A_N$($\uparrow$) & $Diff$($\downarrow$) & $F_N$($\downarrow$) \\ \hline
\multirow{8}{*}{ResNet32} & \emph{Upper-bound} & \emph{72.99} & - & - & 93.59 & - & - \\ \cline{2-8} 
 & EWC & 34.16 & 38.83 & 36.20 & 51.25 & 42.34 & 42.06  \\
 & GEM & 24.93 & 48.06 & 42.04 & 39.06 & 54.53 & 56.47 \\
 & LwF & 16.73 & 56.26 & 51.62 & 36.18 & 57.41 & 59.19 \\
 & DER++ & 31.03 & 41.96 & 22.63 & 45.59 & 48.00 & 47.82 \\
 & HAT & 30.43 & 42.56 & 36.73 & 37.00 & 56.59 & 51.35 \\
 & UCL & 29.77 & 43.22 & 31.99 & 31.71 & 61.88 & 58.89 \\
 & MAS & 29.58 & 43.41 & 22.14 & 49.59 & 44.00 & 33.90  \\ \hline
\multirow{6}{*}{ViT} & \emph{Upper-bound} & \emph{86.07} & - & - & 80.59 & - & - \\ \cline{2-8} 
 & LwF & 71.93 & 14.14 & \underline{6.89} &  19.02 & 61.57 & 44.23 \\
 & MAS & 76.71 & 9.36 & 13.02 & 65.13 & 15.46 & 13.5 \\
 & L2P & 80.63 & 5.44 & 10.93 & 69.06 & 11.59 & \underline{9.53} \\
 & DualPrompt & \underline{83.18} & \underline{2.89} & 8.42 & \underline{70.14} & \underline{10.45} & 10.11 \\
 & Pro-KT &  \textbf{84.07} & \textbf{2.00} & \textbf{5.43} & \textbf{71.70} & \textbf{8.89} & \textbf{5.19} \\ \hline
\end{tabular}
\caption{Results($\%$) of known samples classification on Split CIFAR100 and 5-datasets. For the Split CIFAR100 dataset, we report the results over 10 tasks with 10 classes per task. For the 5-datasets, we report the results over 5 tasks where each dataset refers to a task. We used two backbones (i.e., ResNet32 and ViT) in the experiments.}
\label{tabel_2}
\end{table*}

\myparagraph{Results on Unknown Detection}
Table \ref{table_1} shows the results of unknown detection. From the results, we can see that the baselines perform poorly on the OwCL problem, indicating that these algorithms struggle to maintain satisfactory open detection performance after learning a sequence of tasks. In contrast, Pro-KT consistently outperforms all compared methods across various configurations, as evident from both the $AUC_N$ (increased) and $FPR_N$ (decreased).
Moreover, we examine the performance under different open detection threshold learning strategies. By comparing the $AUC_N$ for Pro-KT without task IDs, we observe a performance drop of the basic Pro-KT when task identifiers are agnostic in the test. This suggests that task-specific knowledge significantly enhances test performance in OwCL, and the utilization of task IDs further refines open detection boundaries.

The superior performance of our Pro-KT, especially in transferring knowledge from open data across tasks, underscores the success of our proposed prompt bank in accumulating and transferring knowledge. Overall, Pro-KT demonstrates substantial improvements in open detection performance, while effectively mitigating the selection of a promising threshold with adaptive strategies.

\begin{figure*}[ht]\centering
    \subfloat[\centering]{
    \label{fig:subfig1}\includegraphics[width=0.23\textwidth]{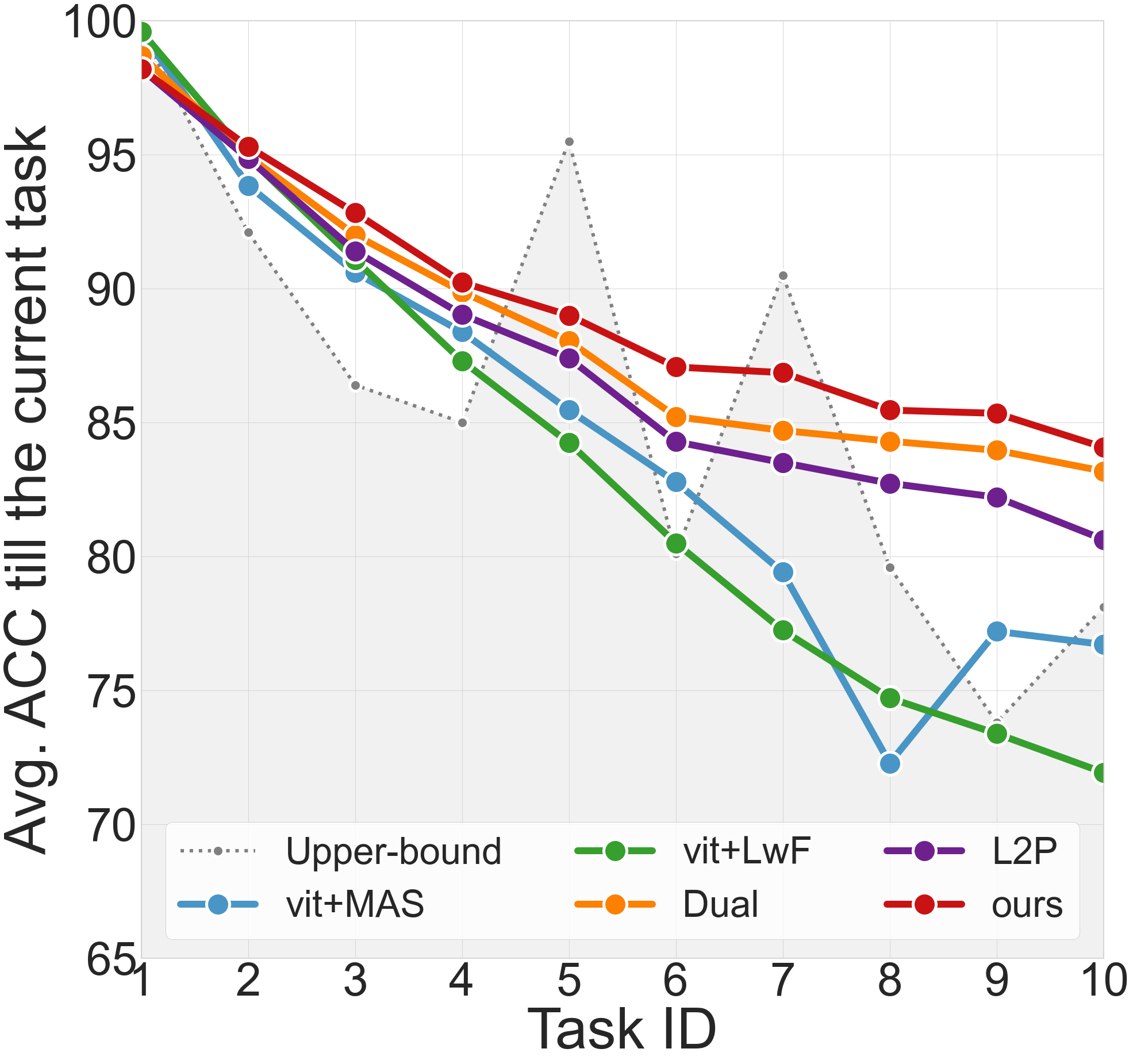}
    }
    \subfloat[\centering]{
    \label{fig:subfig2}\includegraphics[width=0.23\textwidth]{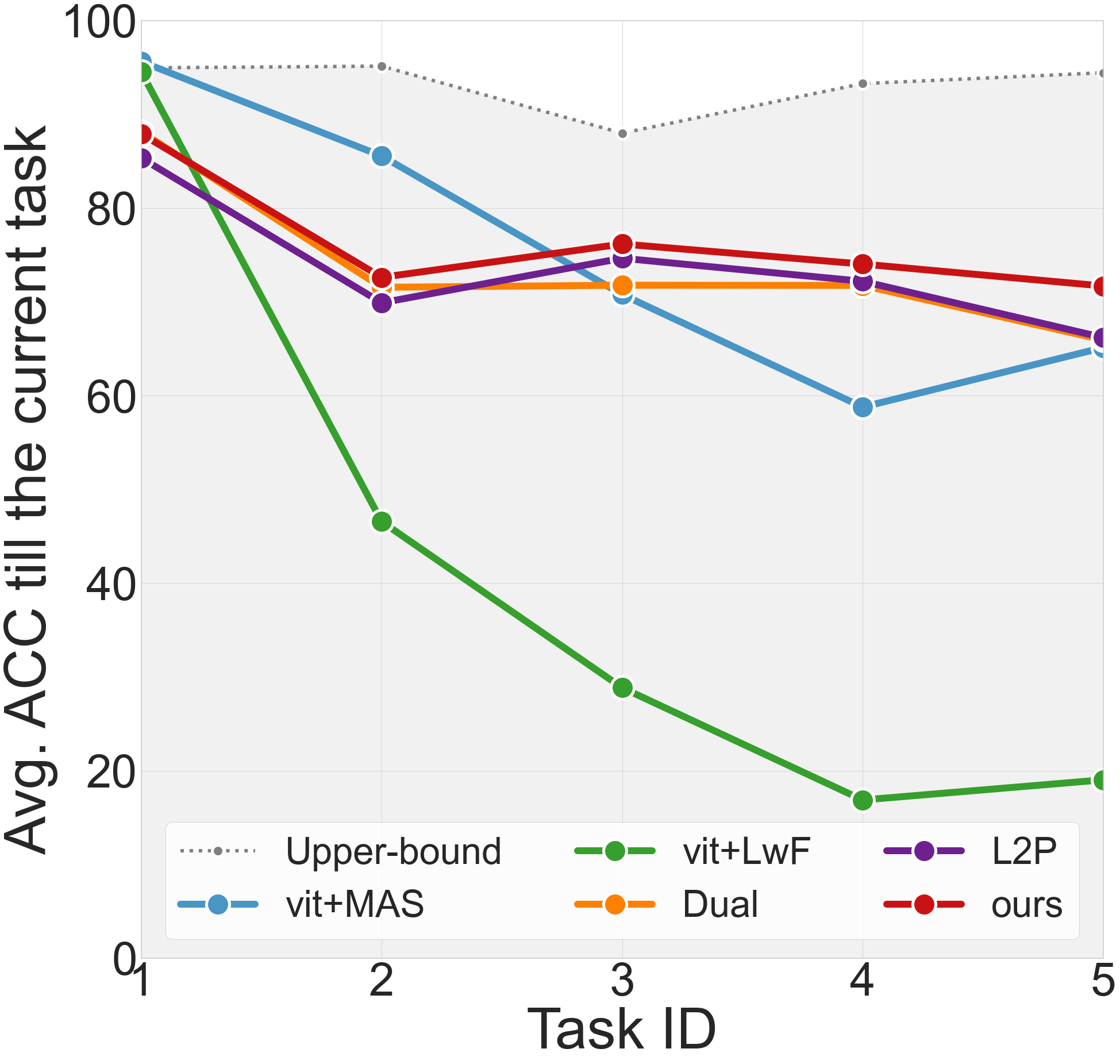}
    }
    \subfloat[\centering]{
    \label{fig:subfig3}\includegraphics[width=0.23\textwidth]{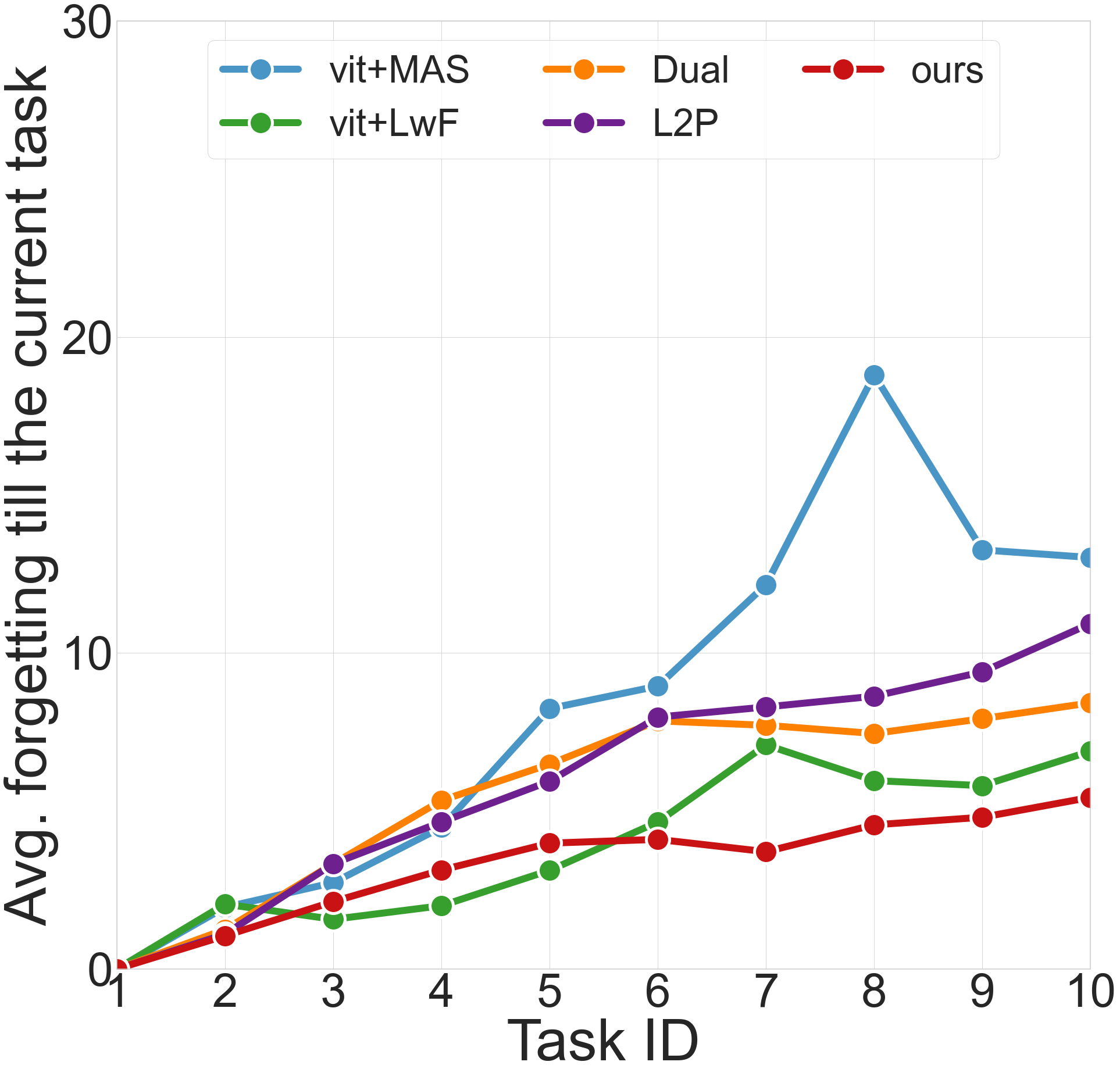}
    }
    \subfloat[\centering]{
    \label{fig:subfig4}\includegraphics[width=0.23\textwidth]{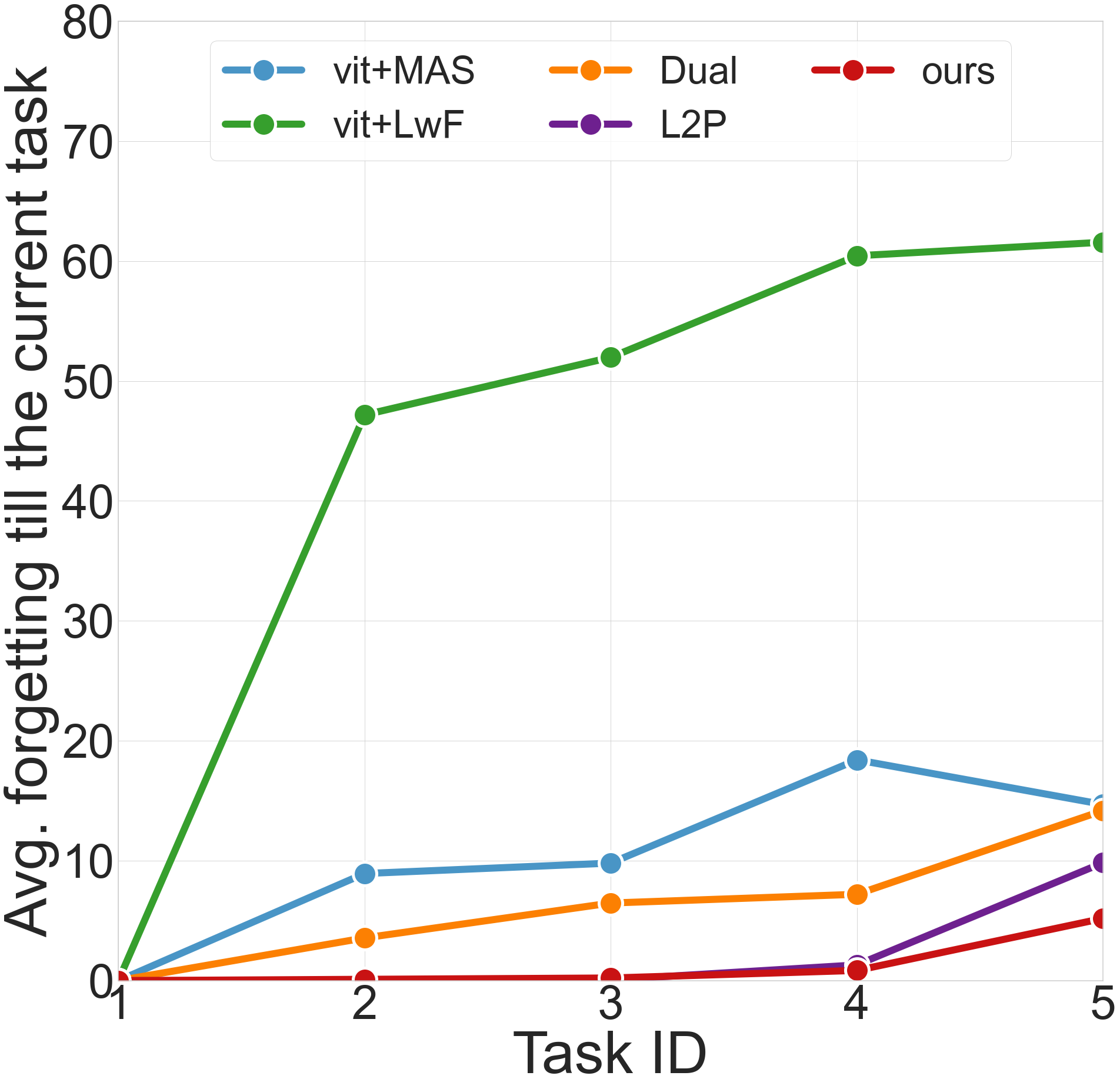}
    }
    \caption{\textbf{(a) and (b)}: Variation Curve of $A_N$ on Split CIFAR100 and 5-datasets. \textbf{(c) and (d)}: Variation Curve of $F_N$ on Split CIFAR100 and 5-datasets.}
    \label{fig_3}
\end{figure*}

\myparagraph{Results on Known Samples Classification}
In this experiment, we evaluate the performance of known classification. 
In particular, Pro-KT does not require task identifiers for the classification of known samples, enabling its application in task-agnostic class-incremental settings. 
We also report the \emph{Upper-bound}, i.e. the offline training of all tasks, which is usually regarded as the upper-bound performance a method can achieve \cite{wang2022learning}. 

Table~\ref{tabel_2} shows the results on the Split CIFAR100 and 5-datasets, respectively. From the results, we can observe that our Pro-KT consistently outperforms other methods and yields SOTA performance. Besides, we observe that Pro-KT achieves higher $A_N$ when ViT is the backbone compared to ResNet32. This is attributed not only to the powerful representation ability of large-scale pre-trained models but also to the adaptability of our prompt learning method for knowledge transfer to pre-trained models. Additionally, a relatively modest performance gap between baseline methods and the upper-bound result has been reported, highlighting the significant performance advantage of our method. It is noted that Pro-KT exhibits the lowest forgetting rate ($F_N$) after learning a sequence of tasks.

\subsection{Additional Results}
\myparagraph{Task Effect} As shown in Fig.~\ref{fig_3}, we illustrate $A_N$ and $F_N$ as line plots to provide further insights into Pro-KT after learning each new task. The results demonstrate the Pro-KT's consistently high performance across tasks and its superiority over baselines with the number of tasks increasing. This demonstrates the effectiveness of our approach in mitigating performance decay and maintaining promising classification results compared to existing benchmarks. Thus, we emphasize the crucial role of knowledge transfer in addressing the OwCL problem.

\begin{table}[hb]
\centering\small
\begin{tabular}{ccccc}
\hline
\multirow{2}{*}{\textbf{\begin{tabular}[c]{@{}c@{}}Unknown \\ Detection\end{tabular}}} & \multicolumn{2}{c}{Split CIFAR100} & \multicolumn{2}{c}{5-datasets} \\ \cline{2-5}
 & $AUC_N$ & $FPR_N$ & $AUC_N$ & $FPR_N$ \\ \hline
w/o task IDs & 91.01 & 41.31 & 82.76 & 50.05 \\
w/o AODB & 50.02 & 87.63 & 46.78 & 88.90 \\
Pro-KT & \textbf{92.69} & \textbf{39.71} & \textbf{88.60} & \textbf{45.70} \\ \hline
\multirow{2}{*}{\textbf{\begin{tabular}[c]{@{}c@{}}Known \\ Classification\end{tabular}}} &\multicolumn{2}{c}{Split CIFAR100} & \multicolumn{2}{c}{5-datasets} \\ \cline{2-5}
 & $A_N$ & $F_N$ & $A_N$ & $F_N$ \\ \hline
w/o prompt bank & 40.49 & 17.21 & 15.42 & 40.19 \\
Pro-KT & \textbf{84.07} & \textbf{5.43} & \textbf{71.70} & \textbf{5.19} \\ \hline
\end{tabular}
\caption{Ablation study on the variants of Pro-KT.}
\label{table_3}
\end{table}

\myparagraph{Ablation Study}
To show the importance of each key component of Pro-KT, we conduct ablation studies on 1) task identifiers in determining the threshold, 2) the whole task-aware open-set boundary, and 3) the prompt bank.
The results are shown in Table~\ref{table_3}. 

First, we remove the task identifiers in determining the threshold. The performance has a small drop (compared with the intact Pro-KT), suggesting that when tasks are diverse, adding task identifiers for the instruction of threshold choosing indeed facilitates knowledge sharing and mitigates interference between dissimilar tasks.
Second, we eliminate the whole open-set boundary. As a result, the model does not have the ability of open detection, where the $AUC_N$ value is only about 50, and the $FPR_N$ is the highest.
Finally, we ablate the entire prompt bank and directly fine-tune the pre-trained backbone and a trainable classifier for each task. The significant decrease in performance suggests that a naive fine-tuning suffers catastrophic forgetting in OwCL.

\begin{figure}[ht]\centering
    \subfloat[\centering\textbf{t-SNE}]{
    \label{fig:subfig5}\includegraphics[width=0.23\textwidth]{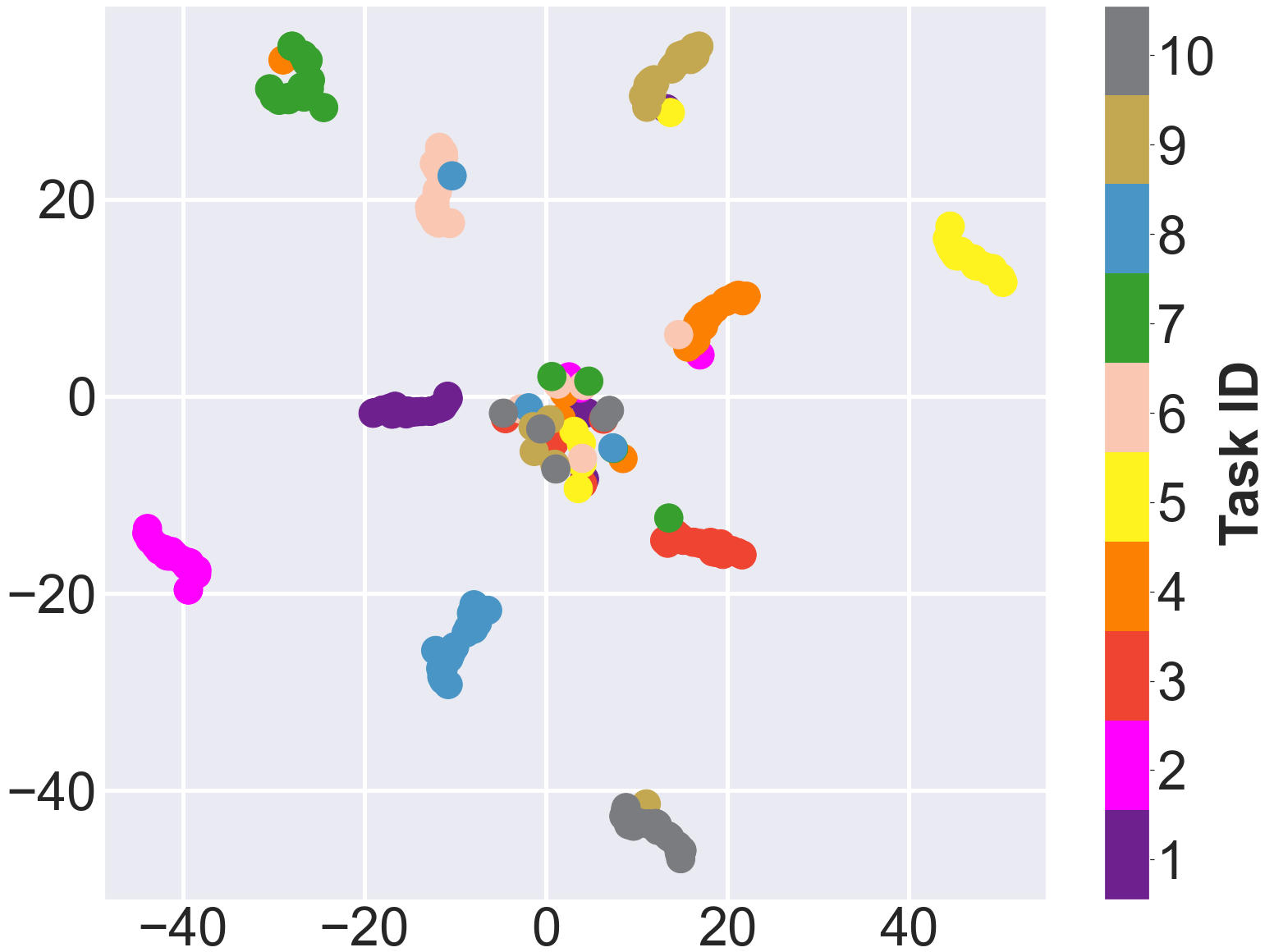}
    }
    \subfloat[\centering\textbf{UMAP}]{
    \label{fig:subfig6}\includegraphics[width=0.23\textwidth]{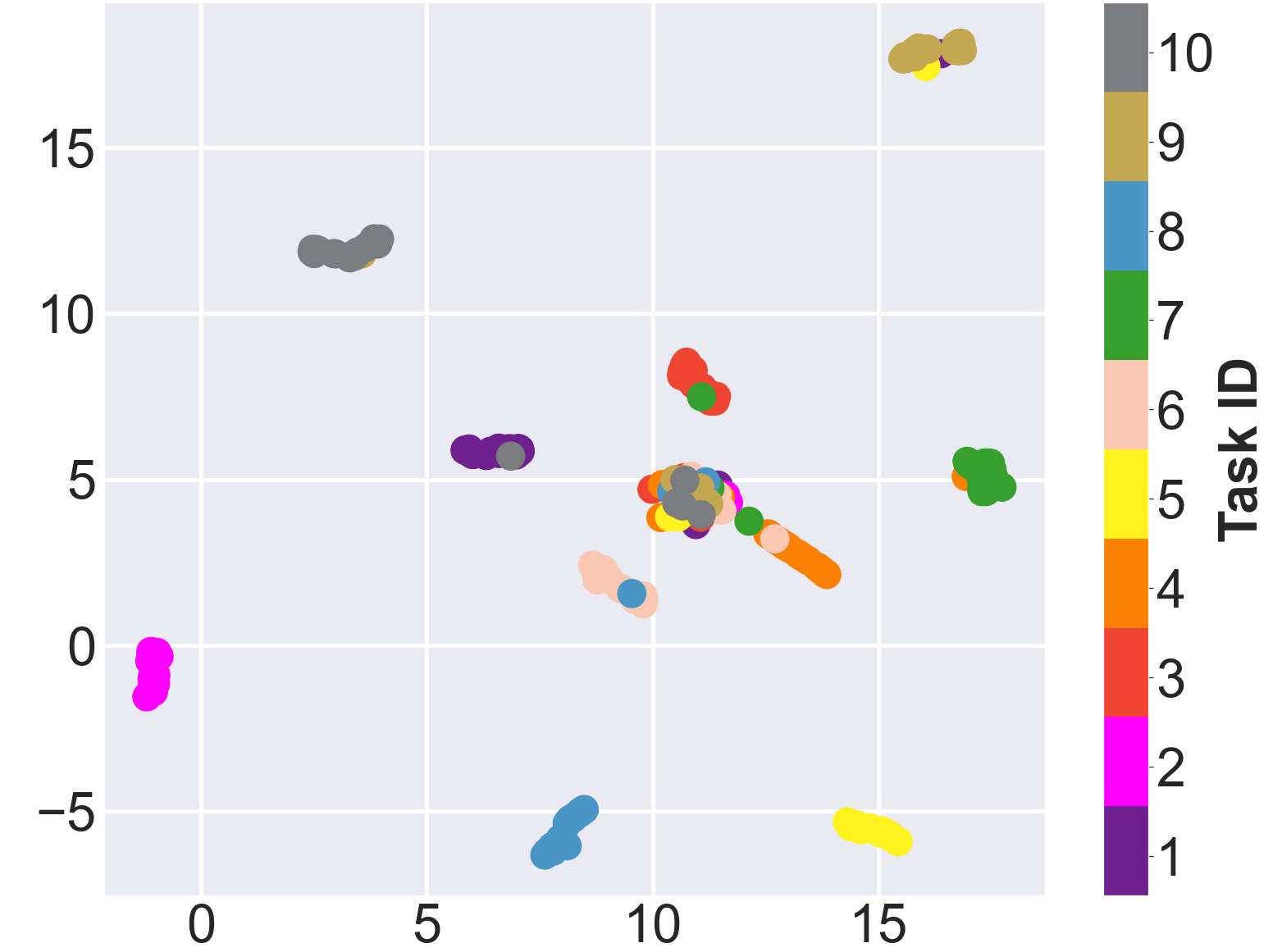}
    }
    \caption{Visualization of the prompt bank via T-SNE and UMAP after training on all tasks (Split CIFAR100 dataset).}
    \label{fig:visualization}
\end{figure}

\myparagraph{Visualization}
To understand the mechanism of knowledge transfer and the different types of knowledge learned within the prompt bank, we visualize all the prompts using t-SNE \cite{van2008visualizing} and UMAP \cite{mcinnes2018umap} on the dataset Split CIFAR100. 

Specifically, we obtain the prompt bank from the final model after training on the sequence of all tasks $T^1, ..., T^n$. The results are shown in Fig.~\ref{fig:visualization}, where each point represents a prompt. We observe that all the prompts are well-separated with different colors, indicating the model has learned task-specific knowledge. Besides, some prompts are automatically gathered in the center and these prompts are learned through different tasks, which indicates that the model has also learned task-generic knowledge. Hence, both task-specific knowledge and task-generic knowledge can be transferred across all tasks with the proposed prompt bank.

\begin{figure}[h]
\centering
\includegraphics[width=\columnwidth]{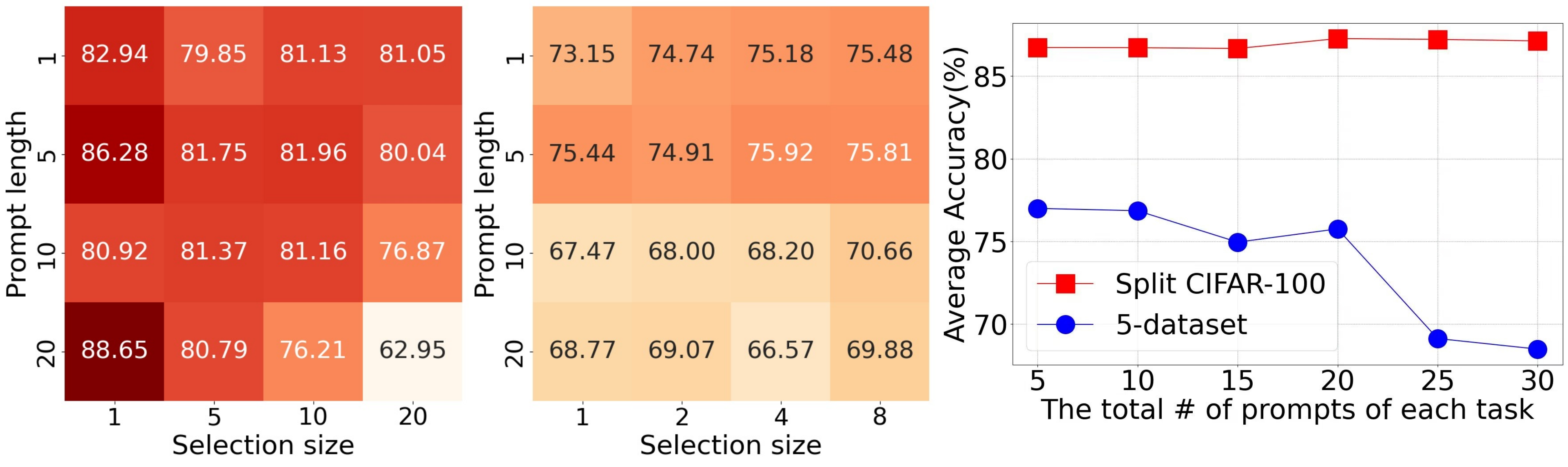}
\caption{\textbf{Left} and \textbf{Middle}: $A_N$ w.r.t prompt length $L_P$ and prompt selection size $K$ on both two datasets with $M=25$. \textbf{Right}: $A_N$ w.r.t. $M$ (i.e., the total number of prompts of each task) with $L_P=5$ and $K=3$.}
\label{fig:parameters}
\end{figure}

\myparagraph{Parameter Sensitivity Analysis}
Recall that there are three key parameters in our Pro-KT, i.e., $M$: the total number of prompts learned from each task, $L_p$: the length of a single prompt, and $K$: the selection size used to prepend the input. Hence, $M \times N$ decides the total size of the prompt bank. Fig.~\ref{fig:parameters} shows the results on two datasets. 
We now analyze the impact of different parameter settings that could affect the proposed Pro-KT:

From the results in Fig.~\ref{fig:parameters} (left-middle), we can see that an oversized selection size $K$ may introduce knowledge under-fitting, while $L_P$ has little effect on performance. For the total number of prompts for each task, we find that a reasonable capacity of the prompt bank is essential for encoding the task-common knowledge and task-specific knowledge. From the results in Fig.~\ref{fig:parameters} (right), it was observed that in the Split CIFAR100 where the 10 tasks are similar, the effect of $M$ on the performance was stable. However, in the 5-datasets where the 5 tasks were dissimilar, increasing $M$ introduced irrelevant prompts, resulting in a decline in performance. Thereby, we safely conclude that when tasks are more dissimilar, the $M$ is more sensitive.

\section{Conclusion}
In this work, we proposed a prompt-based method (called Pro-KT) that achieves knowledge transfer via a novel prompt bank for open-world continual learning. Our Pro-KT involves a unique approach to representing task-specific and task-generic knowledge by attaching the complementary prompts to a pre-trained backbone. 
To determine a task-aware open-set boundary, we design two adaptive threshold-selection strategies, guided by the prompt bank. The experimental results demonstrate the effectiveness of our Pro-KT and its adaptability to different tasks. 

Our future work includes: 1) integrating advanced clustering methods for the fine-grained classification of unknowns, and 2) exploring the forms of knowledge for unknowns and further using the knowledge to help future task learning.

\section{Acknowledgments}
This work was supported by the Natural Science Foundation of Sichuan Province (No. 2022NSFSC0528), the Sichuan Science and Technology Program (No. 2022ZYD0113), Central University Education and Teaching Reforming Program, Southwestern University of Finance and Economics (No. 2023YJG043) and Jiaozi Institute of Fintech Innovation, Southwestern University of Finance and Economics (Nos. kjcgzh20230103, kjcgzh20230201, kjcgzh20230202).
\bibliography{aaai24}

\begin{thebibliography}{36}
\providecommand{\natexlab}[1]{#1}

\bibitem[{Ahn et~al.(2019)Ahn, Cha, Lee, and Moon}]{ahn2019uncertainty}
Ahn, H.; Cha, S.; Lee, D.; and Moon, T. 2019.
\newblock Uncertainty-based continual learning with adaptive regularization.
\newblock In \emph{Advances in Neural Information Processing Systems}, 4392--4402.

\bibitem[{Aljundi et~al.(2018)Aljundi, Babiloni, Elhoseiny, Rohrbach, and Tuytelaars}]{aljundi2018memory}
Aljundi, R.; Babiloni, F.; Elhoseiny, M.; Rohrbach, M.; and Tuytelaars, T. 2018.
\newblock Memory aware synapses: Learning what (not) to forget.
\newblock In \emph{Proceedings of the European Conference on Computer Vision (ECCV)}, 139--154.

\bibitem[{Basart et~al.(2022)Basart, Mantas, Mohammadreza, Jacob, and Dawn}]{basart2022scaling}
Basart, S.; Mantas, M.; Mohammadreza, M.; Jacob, S.; and Dawn, S. 2022.
\newblock Scaling Out-of-Distribution Detection for Real-World Settings.
\newblock In \emph{International Conference on Machine Learning}.

\bibitem[{Bendale and Boult(2015)}]{bendale2015towards}
Bendale, A.; and Boult, T. 2015.
\newblock Towards open world recognition.
\newblock In \emph{Proceedings of the IEEE Conference on Computer Vision and Pattern Recognition}, 1893--1902.

\bibitem[{Buzzega et~al.(2020)Buzzega, Boschini, Porrello, Abati, and Calderara}]{buzzega2020dark}
Buzzega, P.; Boschini, M.; Porrello, A.; Abati, D.; and Calderara, S. 2020.
\newblock Dark experience for general continual learning: a strong, simple baseline.
\newblock \emph{Advances in neural information processing systems}, 33: 15920--15930.

\bibitem[{Chan, Rottmann, and Gottschalk(2021)}]{chan2021entropy}
Chan, R.; Rottmann, M.; and Gottschalk, H. 2021.
\newblock Entropy maximization and meta classification for out-of-distribution detection in semantic segmentation.
\newblock In \emph{Proceedings of the ieee/cvf international conference on computer vision}, 5128--5137.

\bibitem[{Dang et~al.(2019)Dang, Cao, Cui, Pi, and Liu}]{8631004-2019}
Dang, S.; Cao, Z.; Cui, Z.; Pi, Y.; and Liu, N. 2019.
\newblock Open Set Incremental Learning for Automatic Target Recognition.
\newblock \emph{IEEE Transactions on Geoscience and Remote Sensing}, 57(7): 4445--4456.

\bibitem[{Ebrahimi et~al.(2020)Ebrahimi, Meier, Calandra, Darrell, and Rohrbach}]{ebrahimi2020adversarial}
Ebrahimi, S.; Meier, F.; Calandra, R.; Darrell, T.; and Rohrbach, M. 2020.
\newblock Adversarial continual learning.
\newblock In \emph{Computer Vision--ECCV 2020: 16th European Conference, Glasgow, UK, August 23--28, 2020, Proceedings, Part XI 16}, 386--402. Springer.

\bibitem[{Fei and Liu(2016)}]{fei2016breaking}
Fei, G.; and Liu, B. 2016.
\newblock Breaking the closed world assumption in text classification.
\newblock In \emph{Proceedings of the 2016 Conference of the North American Chapter of the Association for Computational Linguistics: Human Language Technologies}, 506--514.

\bibitem[{Gu et~al.(2022)Gu, Han, Liu, and Huang}]{gu2022ppt}
Gu, Y.; Han, X.; Liu, Z.; and Huang, M. 2022.
\newblock PPT: Pre-trained Prompt Tuning for Few-shot Learning.
\newblock In \emph{Proceedings of the 60th Annual Meeting of the Association for Computational Linguistics (Volume 1: Long Papers)}, 8410--8423.

\bibitem[{Guo, Liu, and Zhao(2023)}]{guo2023dealing}
Guo, Y.; Liu, B.; and Zhao, D. 2023.
\newblock Dealing with Cross-Task Class Discrimination in Online Continual Learning.
\newblock In \emph{Proceedings of the IEEE/CVF Conference on Computer Vision and Pattern Recognition}, 11878--11887.

\bibitem[{Hendrycks and Gimpel(2016)}]{hendrycks2016baseline}
Hendrycks, D.; and Gimpel, K. 2016.
\newblock A Baseline for Detecting Misclassified and Out-of-Distribution Examples in Neural Networks.
\newblock In \emph{International Conference on Learning Representations}.

\bibitem[{Joseph et~al.(2021)Joseph, Khan, Khan, and Balasubramanian}]{Joseph_2021_CVPR}
Joseph, K.~J.; Khan, S.; Khan, F.~S.; and Balasubramanian, V.~N. 2021.
\newblock Towards Open World Object Detection.
\newblock In \emph{Proceedings of the IEEE/CVF Conference on Computer Vision and Pattern Recognition (CVPR)}, 5830--5840.

\bibitem[{Ke, Liu, and Huang(2020)}]{ke2020continual}
Ke, Z.; Liu, B.; and Huang, X. 2020.
\newblock Continual learning of a mixed sequence of similar and dissimilar tasks.
\newblock \emph{Advances in Neural Information Processing Systems}, 33: 18493--18504.

\bibitem[{Khattak et~al.(2023)Khattak, Rasheed, Maaz, Khan, and Khan}]{khattak2023maple}
Khattak, M.~U.; Rasheed, H.; Maaz, M.; Khan, S.; and Khan, F.~S. 2023.
\newblock Maple: Multi-modal prompt learning.
\newblock In \emph{Proceedings of the IEEE/CVF Conference on Computer Vision and Pattern Recognition}, 19113--19122.

\bibitem[{Kim et~al.(2023)Kim, Xiao, Konishi, Ke, and Liu}]{kim2023open}
Kim, G.; Xiao, C.; Konishi, T.; Ke, Z.; and Liu, B. 2023.
\newblock Open-World Continual Learning: Unifying Novelty Detection and Continual Learning.
\newblock \emph{arXiv preprint arXiv:2304.10038}.

\bibitem[{Kirkpatrick et~al.(2017)Kirkpatrick, Pascanu, Rabinowitz, Veness, Desjardins, Rusu, Milan, Quan, Ramalho, and GrabskaBarwinska}]{kirkpatrick2017overcoming}
Kirkpatrick, J.; Pascanu, R.; Rabinowitz, N.; Veness, J.; Desjardins, G.; Rusu, A.~A.; Milan, K.; Quan, J.; Ramalho, T.; and GrabskaBarwinska, A. 2017.
\newblock Overcoming catastrophic forgetting in neural networks.
\newblock In \emph{PANS}, volume 114, 3521--3526.

\bibitem[{Krizhevsky, Hinton et~al.(2009)}]{krizhevsky2009learning}
Krizhevsky, A.; Hinton, G.; et~al. 2009.
\newblock Learning multiple layers of features from tiny images.
\newblock \emph{Master's thesis, University of Tront}.

\bibitem[{Lester, Al-Rfou, and Constant(2021)}]{lester2021power}
Lester, B.; Al-Rfou, R.; and Constant, N. 2021.
\newblock The Power of Scale for Parameter-Efficient Prompt Tuning.
\newblock In \emph{Proceedings of the 2021 Conference on Empirical Methods in Natural Language Processing}, 3045--3059.

\bibitem[{Li and Liang(2021)}]{li2021prefix}
Li, X.~L.; and Liang, P. 2021.
\newblock Prefix-Tuning: Optimizing Continuous Prompts for Generation.
\newblock In \emph{Proceedings of the 59th Annual Meeting of the Association for Computational Linguistics and the 11th International Joint Conference on Natural Language Processing (Volume 1: Long Papers)}, 4582--4597.

\bibitem[{Li and Hoiem(2017)}]{li2017learning}
Li, Z.; and Hoiem, D. 2017.
\newblock Learning without forgetting.
\newblock \emph{IEEE Transactions on Pattern Analysis and Machine Intelligence}, 40(12): 2935--2947.

\bibitem[{Liang, Li, and Srikant(2018)}]{liang2018enhancing}
Liang, S.; Li, Y.; and Srikant, R. 2018.
\newblock Enhancing The Reliability of Out-of-distribution Image Detection in Neural Networks.
\newblock In \emph{International Conference on Learning Representations}.

\bibitem[{Liu et~al.(2023{\natexlab{a}})Liu, Mazumder, Robertson, and Grigsby}]{liu2023ai}
Liu, B.; Mazumder, S.; Robertson, E.; and Grigsby, S. 2023{\natexlab{a}}.
\newblock AI Autonomy: Self-Initiation, Adaptation and Continual Learning.
\newblock \emph{AI Magazine}.

\bibitem[{Liu et~al.(2023{\natexlab{b}})Liu, Yuan, Fu, Jiang, Hayashi, and Neubig}]{liu2023pre}
Liu, P.; Yuan, W.; Fu, J.; Jiang, Z.; Hayashi, H.; and Neubig, G. 2023{\natexlab{b}}.
\newblock Pre-train, prompt, and predict: A systematic survey of prompting methods in natural language processing.
\newblock \emph{ACM Computing Surveys}, 55(9): 1--35.

\bibitem[{Liu et~al.(2020)Liu, Wang, Owens, and Li}]{liu2020energy}
Liu, W.; Wang, X.; Owens, J.; and Li, Y. 2020.
\newblock Energy-based out-of-distribution detection.
\newblock \emph{Advances in Neural Information Processing Systems}, 33: 21464--21475.

\bibitem[{Lopez-Paz and Ranzato(2017)}]{lopez2017gradient}
Lopez-Paz, D.; and Ranzato, M. 2017.
\newblock Gradient episodic memory for continual learning.
\newblock \emph{Advances in Neural Information Processing Systems}, 30.

\bibitem[{McInnes et~al.(2018)McInnes, Healy, Saul, and Gro{\ss}berger}]{mcinnes2018umap}
McInnes, L.; Healy, J.; Saul, N.; and Gro{\ss}berger, L. 2018.
\newblock UMAP: Uniform Manifold Approximation and Projection.
\newblock \emph{Journal of Open Source Software}, 3(29).

\bibitem[{Serra et~al.(2018)Serra, Suris, Miron, and Karatzoglou}]{serra2018overcoming}
Serra, J.; Suris, D.; Miron, M.; and Karatzoglou, A. 2018.
\newblock Overcoming catastrophic forgetting with hard attention to the task.
\newblock In \emph{International Conference on Machine Learning}, 4548--4557. PMLR.

\bibitem[{Smith et~al.(2023)Smith, Karlinsky, Gutta, Cascante-Bonilla, Kim, Arbelle, Panda, Feris, and Kira}]{smith2023coda}
Smith, J.~S.; Karlinsky, L.; Gutta, V.; Cascante-Bonilla, P.; Kim, D.; Arbelle, A.; Panda, R.; Feris, R.; and Kira, Z. 2023.
\newblock CODA-Prompt: COntinual Decomposed Attention-based Prompting for Rehearsal-Free Continual Learning.
\newblock In \emph{Proceedings of the IEEE/CVF Conference on Computer Vision and Pattern Recognition}, 11909--11919.

\bibitem[{Van~der Maaten and Hinton(2008)}]{van2008visualizing}
Van~der Maaten, L.; and Hinton, G. 2008.
\newblock Visualizing data using t-SNE.
\newblock \emph{Journal of Machine Learning Research}, 9(11).

\bibitem[{Wang et~al.(2019)Wang, Liu, Wang, Ma, and Yang}]{wang2019forward}
Wang, H.; Liu, B.; Wang, S.; Ma, N.; and Yang, Y. 2019.
\newblock Forward and backward knowledge transfer for sentiment classification.
\newblock In \emph{Asian Conference on Machine Learning}, 457--472. PMLR.

\bibitem[{Wang et~al.(2022{\natexlab{a}})Wang, Zhang, Ebrahimi, Sun, Zhang, Lee, Ren, Su, Perot, Dy et~al.}]{wang2022dualprompt}
Wang, Z.; Zhang, Z.; Ebrahimi, S.; Sun, R.; Zhang, H.; Lee, C.-Y.; Ren, X.; Su, G.; Perot, V.; Dy, J.; et~al. 2022{\natexlab{a}}.
\newblock Dualprompt: Complementary prompting for rehearsal-free continual learning.
\newblock In \emph{Computer Vision--ECCV 2022: 17th European Conference, Tel Aviv, Israel, October 23--27, 2022, Proceedings, Part XXVI}, 631--648. Springer.

\bibitem[{Wang et~al.(2022{\natexlab{b}})Wang, Zhang, Lee, Zhang, Sun, Ren, Su, Perot, Dy, and Pfister}]{wang2022learning}
Wang, Z.; Zhang, Z.; Lee, C.-Y.; Zhang, H.; Sun, R.; Ren, X.; Su, G.; Perot, V.; Dy, J.; and Pfister, T. 2022{\natexlab{b}}.
\newblock Learning to prompt for continual learning.
\newblock In \emph{Proceedings of the IEEE/CVF Conference on Computer Vision and Pattern Recognition}, 139--149.

\bibitem[{Yu and Aizawa(2019)}]{yu2019unsupervised}
Yu, Q.; and Aizawa, K. 2019.
\newblock Unsupervised out-of-distribution detection by maximum classifier discrepancy.
\newblock In \emph{Proceedings of the IEEE/CVF International Conference on Computer Vision}, 9518--9526.

\bibitem[{Zenke, Poole, and Ganguli(2017)}]{pmlr-v70-zenke17a}
Zenke, F.; Poole, B.; and Ganguli, S. 2017.
\newblock Continual Learning Through Synaptic Intelligence.
\newblock In Precup, D.; and Teh, Y.~W., eds., \emph{Proceedings of the 34th International Conference on Machine Learning}, volume~70 of \emph{Proceedings of Machine Learning Research}, 3987--3995. PMLR.

\bibitem[{Zhou et~al.(2022)Zhou, Yang, Loy, and Liu}]{zhou2022learning}
Zhou, K.; Yang, J.; Loy, C.~C.; and Liu, Z. 2022.
\newblock Learning to prompt for vision-language models.
\newblock \emph{International Journal of Computer Vision}, 130(9): 2337--2348.

\end{thebibliography}

\end{document}